\documentclass{article}

\usepackage{amsmath}
\usepackage{amsfonts}
\usepackage{graphicx}
\usepackage{booktabs}
\usepackage{multirow}
\usepackage{placeins}
\usepackage{wrapfig}
\usepackage{float}
\usepackage{capt-of}
\usepackage[dvipsnames,table]{xcolor}
\usepackage{tocloft}

\usepackage[preprint]{corl_2026}

\definecolor{linkpurple}{HTML}{5B006B}
\hypersetup{
  colorlinks=true,
  linkcolor=linkpurple,
  citecolor=linkpurple,
  urlcolor=linkpurple,
  linktoc=all,
}

\definecolor{methodblue}{RGB}{232,244,255}

\title{\textbf{TacCoRL}: Integrating Tactile Feedback into VLA via Simulation}

\author{
  Siyu Ma$^{1,2*}$, Yuqi Liang$^{1, 3*}$, Chang Yu$^{1*\dagger}$,\\
  \textbf{Yunuo Chen$^{1}$, Hao Su$^{2}$, Yixin Zhu$^{4}$, Yin Yang$^{5}$, Chenfanfu Jiang$^{1}$}\\
  {\normalfont $^{1}$University of California, Los Angeles, $^{2}$ University of California, San Diego}\\
  {\normalfont $^{3}$University of Electronic Science and Technology of China}\\
  {\normalfont $^{4}$Peking University, $^{5}$University of Utah}\\
  {\normalfont\footnotesize * Equal contributions. $^\dagger$ Project lead.}
}

\begin{document}

\maketitle
 \vspace{-3.2em}
\begin{abstract}
Vision-language-action (VLA) models provide strong visual, language, and action priors for robot manipulation, but visual observations alone often miss the local contact state required for contact-rich tasks. We present \textbf{TacCoRL}, a scalable framework that injects \textbf{Tac}tile feedback into VLA policies and improves them through sim-real \textbf{Co}-training and simulation-based reinforcement learning (\textbf{RL}), without requiring large-scale tactile pretraining or extensive real-world contact exploration. The key idea is not only adding touch as an input, but learning how contact readings should modulate action responses in near-failure states that are rare in demonstrations and risky to collect on hardware. We use a real-aligned simulator as a closed-loop training environment for contact interaction. Mixed simulated and real trajectories first warm-start tactile-conditioned actions in the pretrained policy. Reinforcement learning with verifiable task rewards then optimizes the policy using simulated contact rollouts. It reinforces tactile-conditioned actions that lead to task completion, while a supervised objective on real trajectories keeps the refined policy anchored to deployment visual, tactile, and action distributions. The resulting policy transfers directly to the real robot without privileged simulation state or online real-world RL. Across four bimanual contact-rich tasks, the final visuo-tactile policy achieves an average success rate of $72.5\%$, compared to baseline of $50.0\%$. Result videos and more details are available at \url{https://tac-corl.github.io/}.
\end{abstract}

\addtocontents{toc}{\protect\setcounter{tocdepth}{-1}}
\vspace{-0.7em}
\begin{figure}[H]
  \centering
  \includegraphics[width=\linewidth]{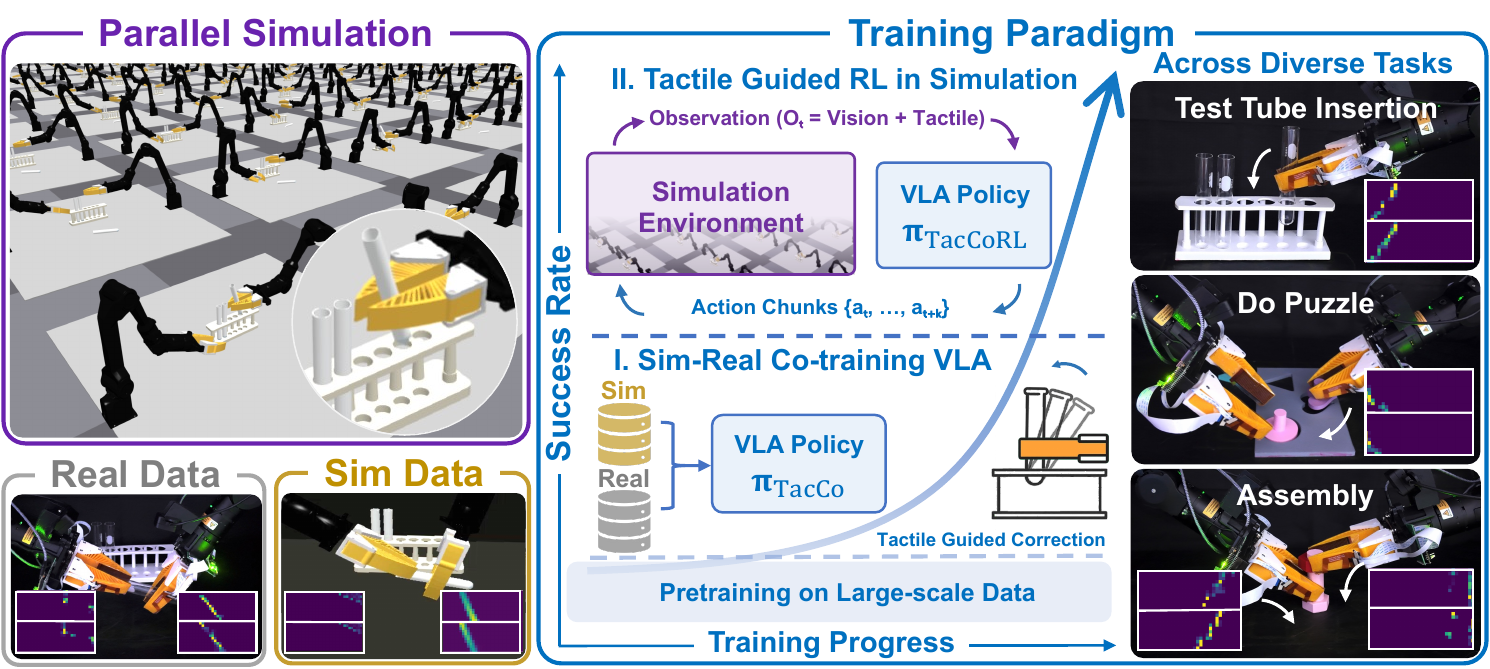}
  \vspace{-0.8em}
    \caption{\textit{Left:} We collect real and simulated visuo-tactile trajectories from aligned real-world and simulation setups.
  \textit{Center:} Sim-real co-training gives the policy an initial tactile-conditioned action prior, and tactile-guided RL in a real-aligned simulator refines closed-loop contact corrections.
  \textit{Right:} We deploy the policy directly to the real world, where it achieves high success rates across diverse contact-rich manipulation tasks.
  }
  \label{fig:teaser}
  \vspace{-0.6em}
\end{figure}

\section{Introduction}
Precise contact-rich manipulation requires vision to interpret the scene and touch to resolve interactions at the contact interface. Vision-Language-Action (VLA) models provide strong visual representations, language grounding, and action priors from large-scale pretraining data \citep{zitkovich2023rt,kim2024openvla,black2024pi_0,intelligence2025pi_}, making them a compelling foundation for robotic manipulation. However, in contact-rich tasks such as insertion, assembly, and in-hand manipulation, cameras often cannot reliably observe whether the manipulated part is properly aligned, where contact occurs, or how pressure is distributed. Tactile sensing \citep{yuan2017gelsight,huang2024vitac,zhao2025touch,huang2025vt} provides this complementary signal by revealing local contact location, pressure distribution, and misalignment. While recent work develops tactile capabilities through large tactile datasets \citep{yang2022touch,cheng2025touch100k,higuerasparsh}, tactile-language-action models \citep{hao2025tla,zhang2026vtla,cheng2025omnivtla,huang2025tactile,bi2025vla,zhang2026tacvla}, or inference-time tactile guidance \citep{zhang2026touchguide}, we investigate whether tactile feedback can instead be \emph{integrated into an existing VLA backbone} without large-scale tactile-labeled pretraining.

Real demonstrations alone are ill-suited for this tactile-aware injection problem. They are costly to scale, emphasize successful nominal behavior, and underrepresent the near-failure cases where tactile feedback is most critical: slight misalignment, contact on the wrong surface, and unstable grasp–object interactions. Imitation learning can expose the policy to tactile observations, but it provides limited supervision on how contact should influence actions in off-nominal states; collecting such cases directly on hardware risks sensor damage and incurs slow and costly resets. We therefore require a safe, resettable, and verifiable simulated tactile environment \citep{xu2023efficient,akinola2025tacsl,huang2025vt,li2026taccel}, where contact conditions can be systematically varied and learned from in closed loop.

We use simulation as the tactile-learning environment, aligned with the real task setup through matched scene configuration, robot controller response, and tactile-contact interfaces calibration \citep{huang2025vt,sha2026efficient}. Mixed sim–real co-training \citep{maddukuri2025sim,lei2026mechanistic} provides the pretrained VLA~\citep{black2024pi_0, intelligence2025pi_} with an initial tactile-conditioned action prior. Static demonstrations cannot teach the policy to recover from its own contact-induced errors. We therefore apply sparse-reward reinforcement learning (RL) in simulation to reinforce successful contact behaviors \citep{tan2025interactive,chen2025pi_,li2025simplevla,shi2026beyond,zhang2026rl}, while a supervised real-data anchor keeps the refined policy aligned with real-world distribution statistics.

\textbf{TacCoRL} treats tactile integration as an efficient sim-to-real post-training framework for pretrained VLA policies (see Figure~\ref{fig:teaser}). Our framework offers three key contributions: (1) a tactile-conditioning interface that adapts pretrained VLA policies without requiring large-scale tactile pretraining or training tactile policies from scratch; (2) a sim-to-real post-training pipeline that combines sim-real co-training with verifiable rewards in simulation, anchored by real data to learn transferable contact-guided action refinement; and (3) real-robot evaluations on four bimanual contact-rich manipulation tasks, demonstrating improved success and robustness over vision- and imitation-only baselines.

\vspace{-1.0em}
\section{Related Work}

\paragraph{Tactile-Augmented VLA Policies.}
Vision-language-action models provide strong pretrained priors, but widely used backbones are still trained mainly on vision, language, and proprioception \citep{zitkovich2023rt,kim2024openvla, black2024pi_0, intelligence2025pi_}. Contact-rich manipulation often depends on fine-grained contact geometry, local contact states, and subtle force variations that are occluded or visually ambiguous, motivating tactile sensing as a complementary modality \citep{yuan2017gelsight,alspach2019soft,huang2024vitac,zhao2025embedding,choi2026wild,li2026simultaneous}. Prior work develops tactile capability through tactile-language-action models \citep{hao2025tla,zhang2026vtla,cheng2025omnivtla}, large tactile datasets \citep{yang2022touch,cheng2025touch100k}, cross-sensor tactile representations \citep{higuerasparsh,xu2025unit,feng2025anytouch}, and sensor-vision-language alignment objectives \citep{radford2021learning,yang2024binding}, or by augmenting existing VLA policies with tactile inputs \citep{zhao2025touch,bi2025vla}, tactile-aware modules \citep{huang2025tactile,zhang2026tacvla,gubernatorov2026hapticvla}, or inference-time guidance \citep{zhang2026touchguide}. Our work follows the augmentation view: it injects tactile feedback into an existing VLA backbone rather than relying on large-scale tactile pretraining or training a tactile policy from scratch.

\vspace{-0.4em}
\paragraph{Sim-Real Post-Training for Contact-Rich Policies.} Post-training refines learned policies beyond demonstrations, which is useful when important contact states are rare in supervised data \citep{zhang2026rl}. Recent VLA work studies online post-training with simulated or real-world rollouts \citep{tan2025interactive,chen2025pi_,xu2026rl}, buffered experience and scalable RL training \citep{li2025simplevla,zang2025rlinf,intelligence2025pi}, offline advantage-based updates \citep{zhang2026balancing}, and sim-real regularization \citep{shi2026beyond}. Related work outside VLAs applies RL to diffusion policies \citep{chi2025diffusion,ren2025diffusion,jiang2025adaptive,zou2026d2ppo}, residual controllers \citep{johannink2019residual,alakuijala2021residual}, shared-autonomy controllers \citep{sha2026efficient}, and simulation-based dexterous manipulation \citep{huang2025vt,fang2026sim,barreiros2026careful}. Our method develops a tactile-aware sim-to-real post-training procedure for adapting existing VLA policies to contact-rich manipulation.

\vspace{-0.6em}
\section{Method}
\label{sec:method}

\begin{figure}[t]
  \centering
  \includegraphics[width=\linewidth]{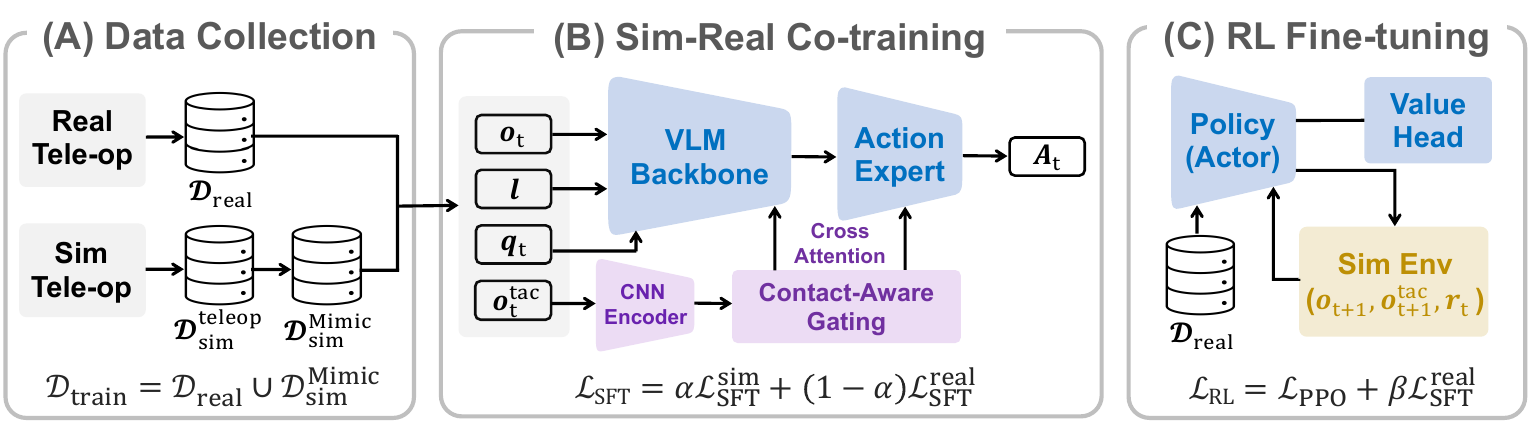}
  \vspace{-0.8em}
  \caption{\textbf{Pipeline.} (A) We collect real demonstrations $\mathcal{D}_{\mathrm{real}}$ together with simulated tele-operation data $\mathcal{D}^{\mathrm{teleop}}_{\mathrm{sim}}$ , and further scale up $\mathcal{D}^{\mathrm{teleop}}_{\mathrm{sim}}$ using MimicGen to obtain $\mathcal{D}^{\mathrm{Mimic}}_{\mathrm{sim}}$. (B) During sim-real co-training, tactile information is encoded and routed through contact-aware gating to modulate both the context of vision-language models (VLM) and the action expert. (C) Interactive simulation rollouts help the tactile-aware policy improve closed-loop task performance, while a supervised real-data loss constrains the update to avoid drifting away from the real robot distribution.}
  \label{fig:pipeline}
  \vspace{-0.3em}
\end{figure}

Our framework adapts a pretrained VLA model to incorporate tactile feedback via a two-stage sim-to-real pipeline. Mixed sim--real co-training first initializes the pretrained VLA with a tactile-conditioned action prior learned from both simulation and real-world expert rollouts. Sparse-reward RL further refines the policy through interactive simulation rollouts, while real-data supervision regularizes the update toward the real-world distribution. At time $t$, the policy receives a language instruction $\boldsymbol{\ell}$, images $\mathbf{o}^v_t$, proprioception $\mathbf{q}_t$, and tactile readings $\mathbf{o}^\tau_t$, and predicts an $H$-step action chunk $\mathbf{A}_t=\mathbf{a}_{t:t+H-1}\in\mathbb{R}^{H\times d_a}$, where $d_a$ denotes the action dimension. We denote the actor observation by
\begin{equation}
    \mathbf{x}_t=(\boldsymbol{\ell},\mathbf{o}^v_t,\mathbf{q}_t,\mathbf{o}^\tau_t),
    \qquad
    \mathbf{A}_t \sim \pi_{\boldsymbol{\theta}}(\cdot \mid \mathbf{x}_t).
    \label{eq:policy_interface}
\end{equation}
The real robot environment is $E_{\mathrm{real}}$, and the simulation environment is $E_{\mathrm{sim}}(\boldsymbol{\psi})$ with alignment parameters $\boldsymbol{\psi}$. Both domains expose $\mathbf{x}_t$; the simulator additionally exposes privileged state $\mathbf{s}_t$ for reward and critic learning. Figure~\ref{fig:pipeline} summarizes the overall pipeline.

\subsection{Simulation Environment Alignment}
\label{sec:method_sim}
Our method uses simulation as a closed-loop contact environment, rather than solely as an offline data source. To make simulation rollouts effective for policy learning, we align the policy-facing interfaces between simulation and the real-world setup: scene configuration and camera define the visual task context; controller response determines how VLA action chunks change the robot state and affects transfer \citep{li2024evaluating,bronars2026tune}; and tactile statistics determine how contact occurrence, location, magnitude, and temporal variation are presented to the policy \citep{huang2025vt,akinola2025tacsl,li2026taccel}. Details of controller identification, tactile-signal alignment, and camera alignment are provided in Appendix~\ref{app:real_to_sim_to_real}.

\subsection{Tactile-Augmented VLA Policy}
\label{sec:method_tactile_vla}

We instantiate the policy using a $\pi_{0.5}$-style VLA model \citep{intelligence2025pi_}: a strong vision-language backbone provides semantic and visual context, while an action expert generates action chunks via a flow policy \citep{black2024pi_0,intelligence2025pi_}. The base VLA tokenizes language, images, and proprioception into
\begin{equation}
    \mathbf{Z}^{\mathrm{base}}_t=[\mathbf{Z}^\ell,\mathbf{Z}^v_t,\mathbf{Z}^q_t].
    \label{eq:base_tokens}
\end{equation}
Our tactile interface adds a tactile encoder $E_\tau$ and a projection $\mathbf{W}_\tau$ that produce $M$ tactile tokens of hidden size $d$. Motivated by OpenTouch’s contact-history analysis \citep{song2025opentouch}, $E_\tau$ encodes a recent tactile window $\mathbf{h}^\tau_t=\mathbf{o}^\tau_{t-L+1:t}\in\mathbb{R}^{L\times K}$ with history length $L$ and $K$ taxels, capturing how contact is loaded, released, and evolves over time:
\begin{equation}
    \mathbf{Z}^\tau_t = \mathbf{W}_\tau E_\tau(\mathbf{h}^\tau_t) \in \mathbb{R}^{M\times d}.
    \label{eq:tactile_tokens}
\end{equation}
A binary contact gate suppresses the influence of tactile tokens when it contains only background noise. For contact-aware tactile fusion \citep{zhang2026tacvla}, we compute the gate over taxel readings $f_{r,k}$ with activation threshold $\lambda_f$ and active-count threshold $m$:
\begin{equation}
    c_t =
    I\left[
    \max_{r\in[t-L+1,t]}
    \sum_{k=1}^{K}
    I(|f_{r,k}|>\lambda_f)
    \ge m
    \right],
    \qquad
    \widetilde{\mathbf{Z}}^\tau_t=\operatorname{Gate}(\mathbf{Z}^\tau_t,c_t).
    \label{eq:contact_indicator}
\end{equation}
The gated tactile tokens enter the model through two conditioning paths:
\begin{equation}
    \bar{\mathbf{Z}}^{\mathrm{base}}_t
    =
    \mathrm{CrossAttn}(\mathbf{Z}^{\mathrm{base}}_t,\widetilde{\mathbf{Z}}^\tau_t),
    \qquad
    \mathbf{Z}_t=[\bar{\mathbf{Z}}^{\mathrm{base}}_t,\widetilde{\mathbf{Z}}^\tau_t].
    \label{eq:gated_fusion}
\end{equation}
Here, $\operatorname{Gate}(\mathbf{Z}^\tau_t,c_t)$ returns $\mathbf{Z}^\tau_t$ under contact and removes tactile tokens from attention otherwise. The first path allows tactile history to update the vision-language-proprioception context; the second keeps tactile tokens available as conditioning tokens for the action expert at each denoising step. This design maintains the pretrained VLA behavior outside contact while allowing tactile evidence to reshape the action chunk during contact.

\begin{figure}[t]
  \centering
  \includegraphics[width=\linewidth]{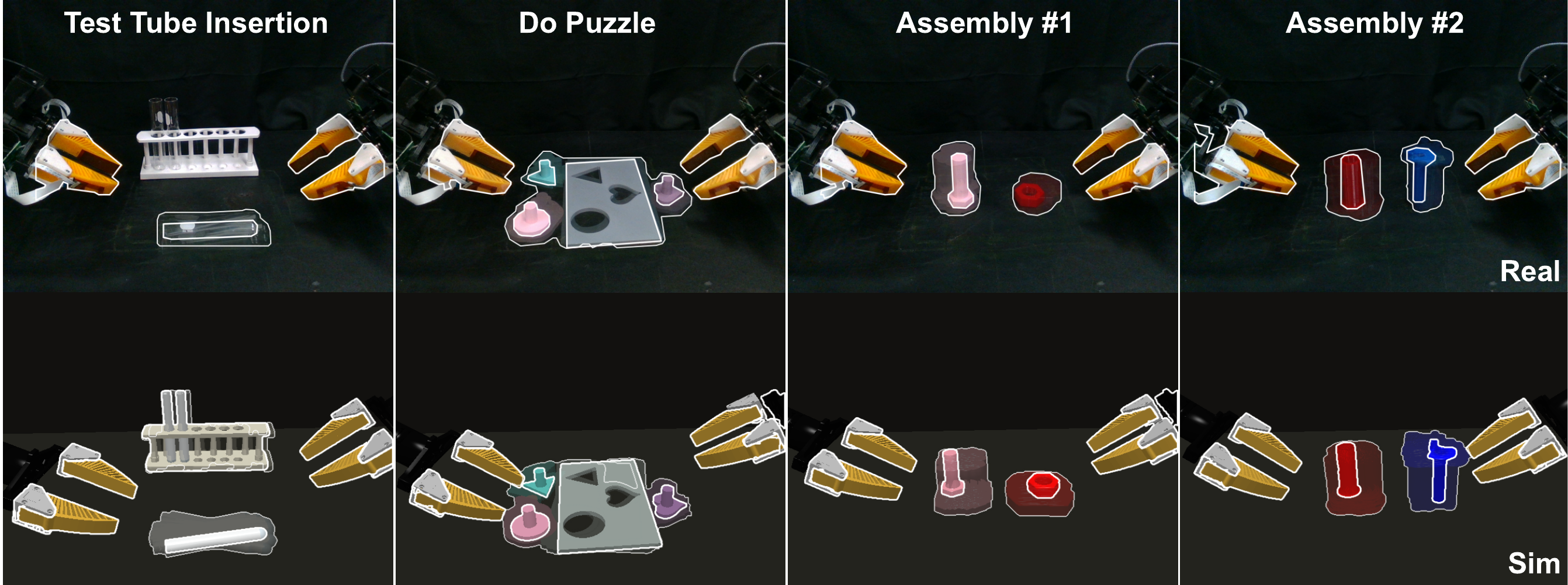}
  \vspace{-0.2em}
  \caption{\textbf{Experimental task settings.} Real and calibrated simulation workspaces for four contact-rich bimanual tasks. The accumulated object placements indicate the pose ranges used for domain randomization and evaluation.}
  \label{fig:exp_setting}
\end{figure}

\subsection{Sim-Real Co-Training}
\label{sec:method_cotrain}

The tactile-augmented VLA provides an interface for touch, but the pretrained backbone has not learned how tactile events should influence actions. We first warm-start the policy via mixed supervised co-training. Let
\begin{equation}
    \mathcal{D}_{\rho}
    =
    \{(\mathbf{x}^i_t,\mathbf{A}^{i,*}_t)\}_{i,t},
    \qquad
    \rho\in\{\mathrm{real},\mathrm{sim}\},
    \label{eq:datasets}
\end{equation}
where both datasets share the same representation $\mathbf{x}_t$. In simulation, we first collect human teleoperation demonstrations $\mathcal{D}^{\mathrm{teleop}}_{\mathrm{sim}}$, then use MimicGen \citep{mandlekar2023mimicgen} to synthesize additional demonstrations $\mathcal{D}^{\mathrm{Mimic}}_{\mathrm{sim}}$ under varied initial object configurations and filter trajectories using the same success predicates as in evaluation; we set $\mathcal{D}_{\mathrm{sim}}=\mathcal{D}^{\mathrm{teleop}}_{\mathrm{sim}}\cup\mathcal{D}^{\mathrm{Mimic}}_{\mathrm{sim}}$. We write $\ell_{\mathrm{flow}}(\boldsymbol{\theta};\mathbf{x},\mathbf{A}^*)$ for the standard conditional flow-matching loss on demonstrated action chunk $\mathbf{A}^*$ conditioned on observation $\mathbf{x}$. The co-training objective is
\begin{equation}
    \mathcal{L}_{\mathrm{co}}(\boldsymbol{\theta})
    =
    \alpha
    \mathrm{E}_{(\mathbf{x},\mathbf{A}^*)\sim\mathcal{D}_{\mathrm{sim}}}
    [\ell_{\mathrm{flow}}(\boldsymbol{\theta};\mathbf{x},\mathbf{A}^*)]
    +
    (1-\alpha)
    \mathrm{E}_{(\mathbf{x},\mathbf{A}^*)\sim\mathcal{D}_{\mathrm{real}}}
    [\ell_{\mathrm{flow}}(\boldsymbol{\theta};\mathbf{x},\mathbf{A}^*)].
    \label{eq:cotrain_loss}
\end{equation}
The coefficient $\alpha\in[0,1]$ controls the relative contribution of simulated data. This ratio is not merely a data-balance hyperparameter: effective co-training should preserve task and contact structure across domains without diluting real-only deployment cues \citep{lei2026mechanistic}.

Co-training has two practical effects that it provides the policy with a tactile-conditioned prior grounded in real observations, and it offers sparse-reward simulator RL a viable initialization for refining contact behavior. However, this co-training stage remains fixed-dataset imitation that it aligns tactile observations with expert action chunks but provides no on-policy contact rollouts or task-level feedback to improve the policy.

\subsection{Post-Training with a Real-Data Anchor}
\label{sec:method_posttraining}

Post-training then optimizes the policy using the on-policy simulation rollouts. We use standard PPO \citep{schulman2017proximal} with task-level success and failure predicates as sparse rewards, so the update favors rollouts that complete the task under contact-rich interaction. To prevent simulator-specific drift, we optimize RL jointly with a supervised real-data anchor \citep{goecks2019integrating}:
\begin{equation}
    \min_{\boldsymbol{\theta},\boldsymbol{\omega}}
    \mathcal{L}_{\mathrm{RL}}(\boldsymbol{\theta},\boldsymbol{\omega})
    =
    \mathcal{L}_{\mathrm{PPO}}(\boldsymbol{\theta},\boldsymbol{\omega};E_{\mathrm{sim}}(\boldsymbol{\psi}))
    +
    \beta
    \mathrm{E}_{(\mathbf{x},\mathbf{A}^*)\sim\mathcal{D}_{\mathrm{real}}}
    [\ell_{\mathrm{flow}}(\boldsymbol{\theta};\mathbf{x},\mathbf{A}^*)].
    \label{eq:taccorl_loss}
\end{equation}
Here $\boldsymbol{\omega}$ are the critic parameters, $\mathcal{L}_{\mathrm{PPO}}$ is the clipped PPO loss, and $\beta$ weights the real-data behavior cloning loss.
This stage collects rollouts from the current tactile-conditioned policy and updates it using sparse task-level rewards. The real-data term anchors the policy to real observations and actions, constraining simulator-specific exploitation \citep{fujimoto2021minimalist, shi2026beyond}. After post-training, we discard the critic, reward, and privileged state and deploy the final policy directly to real robots. Appendix~\ref{app:cotrain_training_params} reports co-training and simulator post-training parameters.

\begin{figure}[t]
  \centering
  \begin{minipage}[t]{0.64\linewidth}
    \centering
    \includegraphics[width=\linewidth,trim=0 0 0 3bp,clip]{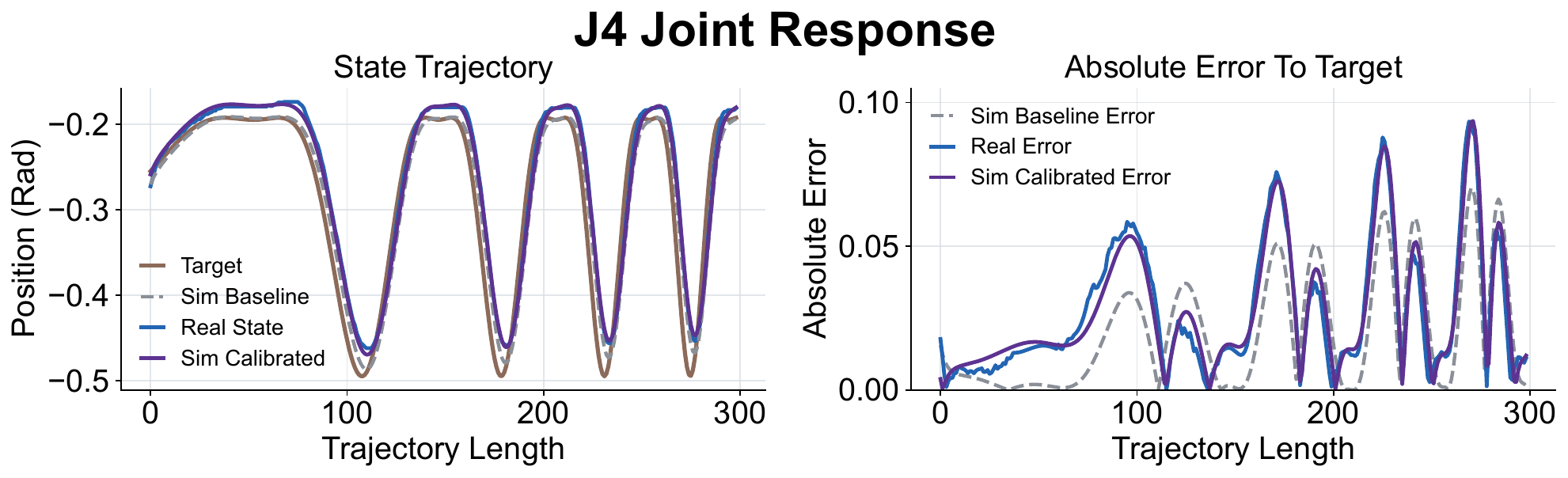}
    \par\vspace{0.05em}
    {\small (A) Controller Response}
  \end{minipage}
  \hfill
  \begin{minipage}[t]{0.35\linewidth}
    \centering
    \includegraphics[width=\linewidth]{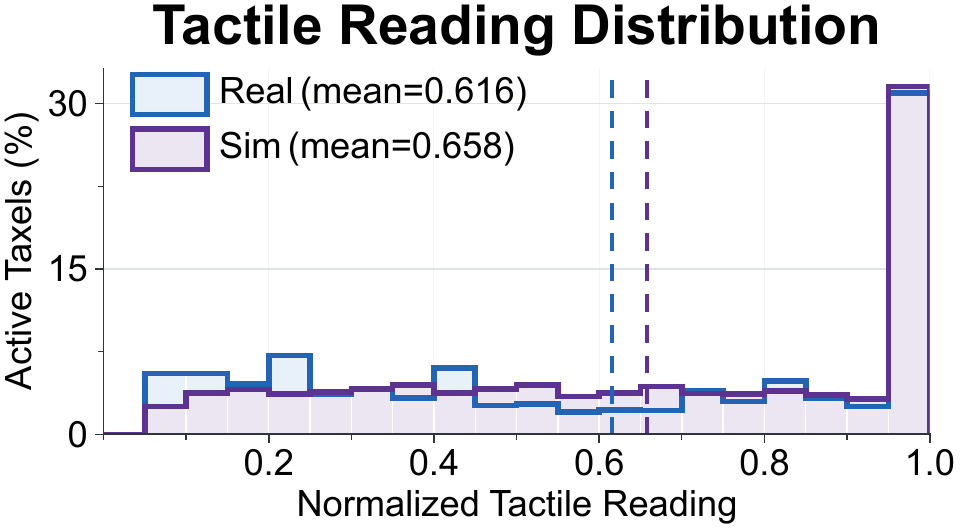}
    \par\vspace{0.05em}
    {\small (B) Tactile Statistics}
  \end{minipage}
  \vspace{-0.2em}
  \caption{\textbf{Controller and tactile alignment.}
  (A) Held-out J4 joint-response replay comparing the target, real and simulated responses before and after controller SysID. (B) Normalized tactile-reading histogram from matched contact rollouts after tactile calibration.
  }
  \label{fig:controller_tactile_calibration}
  \vspace{-0.5em}
\end{figure}

\section{Experimental Results}
\label{sec:experiments}

Our experiments address four related questions: (1) whether the simulated controller responses and tactile readings are sufficiently aligned with those of the real to support closed-loop contact learning (Sec.~\ref{sec:exp_calibration});  (2) whether sim-real co-training provides a strong initialization for simulator RL, and whether tactile feedback further improves the refined policy in simulation (Sec.~\ref{sec:exp_sim_training});  (3) whether policies trained with our method achieve higher success rates on the real (Sec.~\ref{sec:exp_real_quant}); and (4) how the co-training and real-data anchoring weights affect training and real-world deployment (Sec.~\ref{sec:exp_ablation}).

\subsection{Controller and Tactile Alignment}
\label{sec:exp_calibration}

We first validate the two policy-facing interfaces for closed-loop contact learning: controller response and tactile readings. For controller system identification (SysID), we collect short single-joint sweeps on the real robot, replay the same target trajectories in simulation, and identify each joint's proportional gain $K_p$, derivative gain $K_d$, and feed-forward torque $T_{\mathrm{ref}}$ to match the real target-tracking errors. For tactile calibration, we replay matched contact trajectories and compare normalized active-taxel readings across domains.

Figure~\ref{fig:controller_tactile_calibration} summarizes the alignment of the two policy-facing interfaces. Panel (A) shows a representative held-out J4 replay, selected because gravity-compensation mismatch produces the largest pre- versus post-calibration gap. After controller SysID, the calibrated simulator closely matches the real trajectory and tracking error under the same command. Panel (B) shows similar normalized tactile-reading distributions from matched real and simulated contact rollouts after tactile calibration. Together, these results indicate that the action-execution and contact-observation interfaces are sufficiently aligned to support subsequent simulator post-training. Appendices~\ref{app:controller_sysid} and~\ref{app:tactile_alignment} provide additional details.

\begin{figure} [t]
  \centering
  \includegraphics[width=\linewidth]{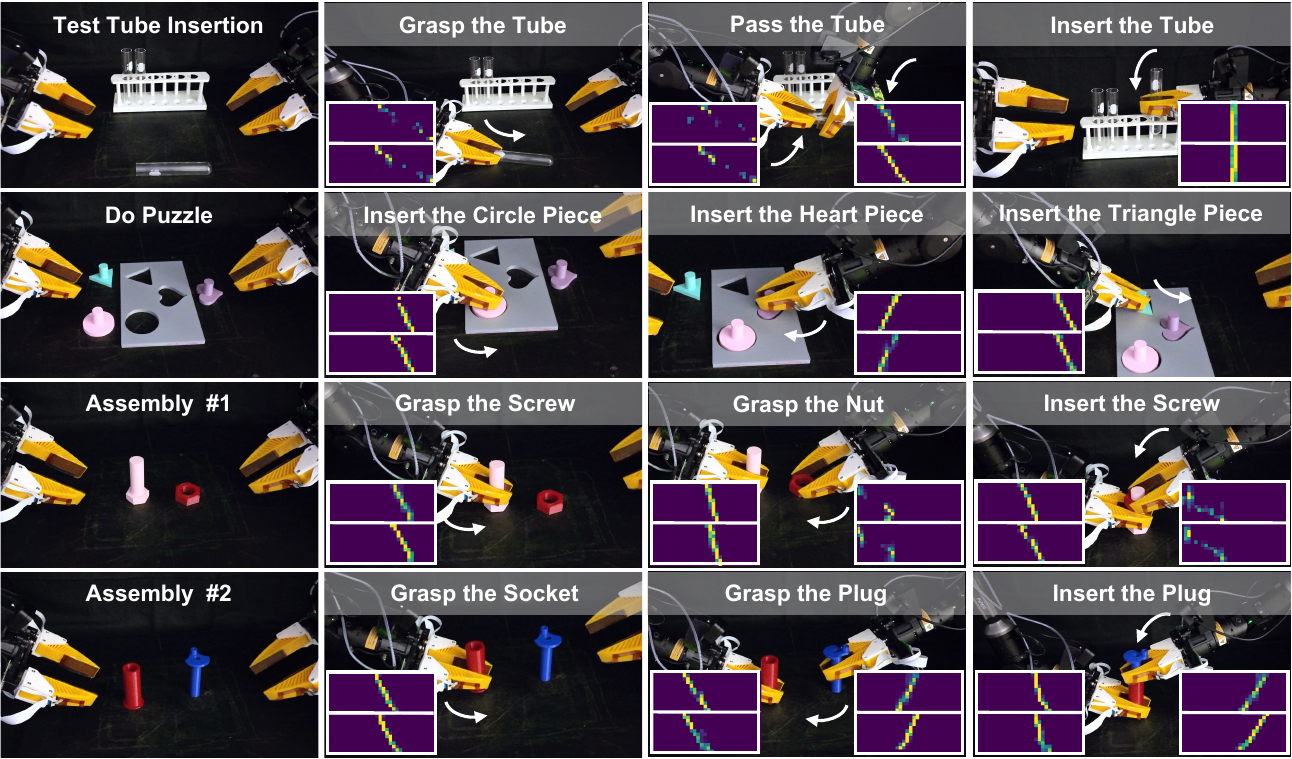}
  \vspace{-0.8em}
  \caption{\textbf{Real-world policy rollouts.} Representative real-robot executions of our post-trained visuo-tactile VLA policy across four contact-rich bimanual tasks.}
  \label{fig:result}
  \vspace{-0.5em}
\end{figure}

\subsection{Task Settings}
\label{sec:exp_setting}
We evaluate on the bimanual platform described in Appendix~\ref{app:robot_setup}, using matched camera, action, and tactile interfaces across the real and simulated setups. Figure~\ref{fig:exp_setting} shows the four contact-rich bimanual tasks and the object-pose randomization used for simulation training; Appendix~\ref{app:domain_randomization} provides the full randomization details. For each task, we collect 50 real demonstrations and 20 simulated teleoperation trajectories, expand the simulated set to 200 trajectories using MimicGen, and evaluate with 128 parallel simulation environments and 20 real-world trials per task.
\subsection{Sim-Real Co-Training and Simulator RL Post-Training}
\label{sec:exp_sim_training}

\begin{figure}[t]
  \centering
  \includegraphics[width=\linewidth]{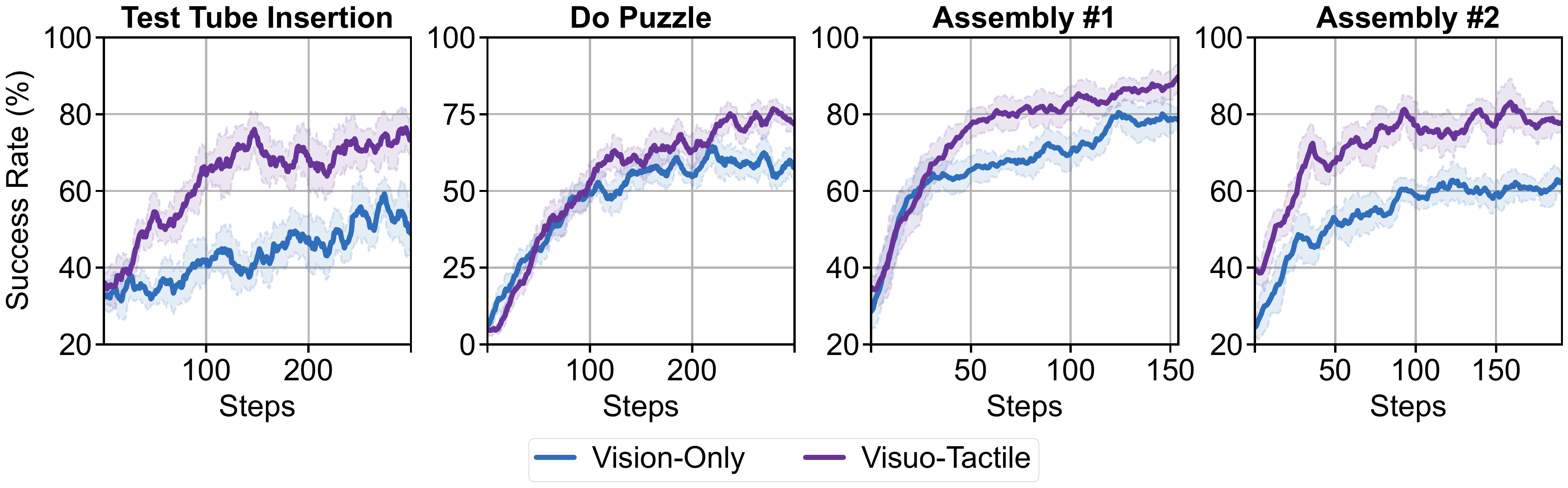}
  \vspace{-0.9em}
  \caption{
\textbf{Tactile feedback improves simulator RL.}
Across all tasks, visuo-tactile policies consistently achieve higher success rates than vision-only policies during simulator RL.
}
  \label{fig:rl_curves}
  \captionof{table}{
  \textbf{Simulation success rates.}
Direct RL from the base VLA model fails across all tasks. Sim-real co-training provides a strong initialization and substantially improves the success rate for simulator-based RL. Incorporating tactile feedback further improves policy performance.
}
  \resizebox{1\textwidth}{!}
  {
  \begin{tabular}{l|cccc|cccc}
      \multicolumn{9}{c}{Simulation Bimanual Setup} \\ \hline
      \multirow{2}{*}{\textbf{Settings}}
        & \multicolumn{4}{c}{\textbf{Vision-Only Policy} (Sim)} \vline
        & \multicolumn{4}{c}{\textbf{Visuo-Tactile Policy} (Sim)} \\
        & \shortstack{Test Tube\\Insertion}
        & \shortstack{Do\\Puzzle}
        & \shortstack{Assembly\\\#1}
        & \shortstack{Assembly\\\#2}
        & \shortstack{Test Tube\\Insertion}
        & \shortstack{Do\\Puzzle}
        & \shortstack{Assembly\\\#1}
        & \shortstack{Assembly\\\#2} \\ \hline
      After Co-Training       & 0.42 & 0.12 & 0.39 & 0.39 & 0.41 & 0.16 & 0.47 & 0.58 \\
      RL Start with Exploration Noise & 0.35 & 0.05 & 0.25 & 0.25 & 0.36 & 0.04 & 0.35 & 0.42 \\
      RL from Base VLA        & 0.00 & 0.00 & 0.00 & 0.00 & 0.00 & 0.00 & 0.00 & 0.00 \\
      RL with Co-Training     & \underline{0.50} & \underline{0.54} & \underline{0.77} & \underline{0.61} & \textbf{0.72} & \textbf{0.71} & \textbf{0.92} & \textbf{0.79}
  \end{tabular}
  }
  \label{tab:sim_rl_posttraining}
  \vspace{-0.4em}
\end{figure}

We evaluate whether sim-real co-training provides a viable initialization for simulator RL and whether tactile feedback further improves the refined policy with RL. Figure~\ref{fig:rl_curves} shows that RL-trained visuo-tactile policies achieve higher final success than vision-only policies across all tasks. This indicates that tactile histories remain useful during on-policy refinement, rather than serving solely as an additional imitation-learning input.

Table~\ref{tab:sim_rl_posttraining} reports RL post-training results for both the co-trained only policy and the same policy evaluated with the exploration noise used at the start of RL. 
Adding exploration noise lowers the initial average success from $33.0\%$ to $22.5\%$ for vision-only policies and from $40.5\%$ to $29.3\%$ for visuo-tactile policies, but this stochasticity is necessary for sparse-reward exploration. In contrast, direct RL from the base VLA remains at $0.0\%$ across all tasks, indicating that sparse rewards alone cannot bootstrap the task sequence or precise contact corrections.

Sim-real co-training therefore changes the post-training problem from exploration from scratch to policy refinement. After RL, average simulator success increases to $60.5\%$ for vision-only policies and $78.5\%$ for visuo-tactile policies. The tactile gains are especially clear in the \emph{Test Tube Insertion} and \emph{Assembly~\#2} tasks, where angular error, blockage, and residual misalignment are often visually occluded but remain reflected in tactile histories.

\vspace{-0.6em}
\vspace{-0.2em}

\subsection{Real-World Evaluation}
\label{sec:exp_real_quant}
\label{sec:exp_real_rollout}

We then evaluate the extent to which tactile feedback, sim-real co-training, and simulator RL improve policy performance on the real robot.

\begin{table}[b]
    \caption{\textbf{Real-world success rates.}
    Rows compare training stages; columns compare vision-only and visuo-tactile policies. Tactile feedback and RL post-training together achieve the highest real-world success rates.}
    \centering
    \resizebox{\textwidth}{!}{%
      \begin{tabular}{@{}c@{}}
        \begin{tabular}{l|cccc|cccc}
          \multicolumn{9}{c}{Real-World Bimanual Setup} \\ \hline
          \multirow{2}{*}{\textbf{Settings}}
            & \multicolumn{4}{c}{\textbf{Vision-Only Policy} (Real)} \vline
            & \multicolumn{4}{c}{\textbf{Visuo-Tactile Policy} (Real)} \\
            & \shortstack{Test Tube\\Insertion}
            & \shortstack{Do\\Puzzle}
            & \shortstack{Assembly\\\#1}
            & \shortstack{Assembly\\\#2}
            & \shortstack{Test Tube\\Insertion}
            & \shortstack{Do\\Puzzle}
            & \shortstack{Assembly\\\#1}
            & \shortstack{Assembly\\\#2} \\ \hline
          Real-Only Fine-Tuning & 0.20 & 0.05 & 0.35 & 0.25 & 0.45 & 0.15 & 0.35 & 0.40 \\
          Sim-Real Co-Training  & 0.35 & 0.10 & 0.40 & 0.35 & 0.50 & 0.25 & 0.45 & 0.55 \\
          RL Post-Training    & \underline{0.35} & \underline{0.25} & \underline{0.80} & \underline{0.60} & \textbf{0.70} & \textbf{0.45} & \textbf{0.95} & \textbf{0.80}
        \end{tabular}
      \end{tabular}%
    }
    \label{tab:real_main}
    \vspace{-15pt}
\end{table}

Figure~\ref{fig:result} qualitatively illustrates the sim-to-real transfer of the learned contact behaviors. In these rollouts, the held object or gripper often occludes the target contact region; the post-trained policy uses tactile feedback to correct local contact errors through incremental translations and reorientations. Quantitatively, tactile feedback increases the average success of RL post-trained policies from $50.0\%$ to $72.5\%$ across the four tasks, indicating that the gain comes from contact sensing rather than from simulator RL alone. \emph{Do Puzzle} is the longest-horizon task because it requires all three pieces to be placed successfully. Real-only fine-tuning and sim-real co-training often stop after partial completion, whereas RL post-training reinforces complete trajectories and improves visuo-tactile success to $45\%$, compared with $15\%$ for real-only fine-tuning and $25\%$ for sim-real co-training, as shown in Table~\ref{tab:real_main}. Appendix~\ref{app:failure_analysis} provides further analysis of baselines that fail during these occluded contact phases.

\subsection{Ablation of Co-Training and Real-Data Anchoring Weights}
\label{sec:exp_ablation}
\begin{figure}[t]
  \centering
  \includegraphics[width=\linewidth]{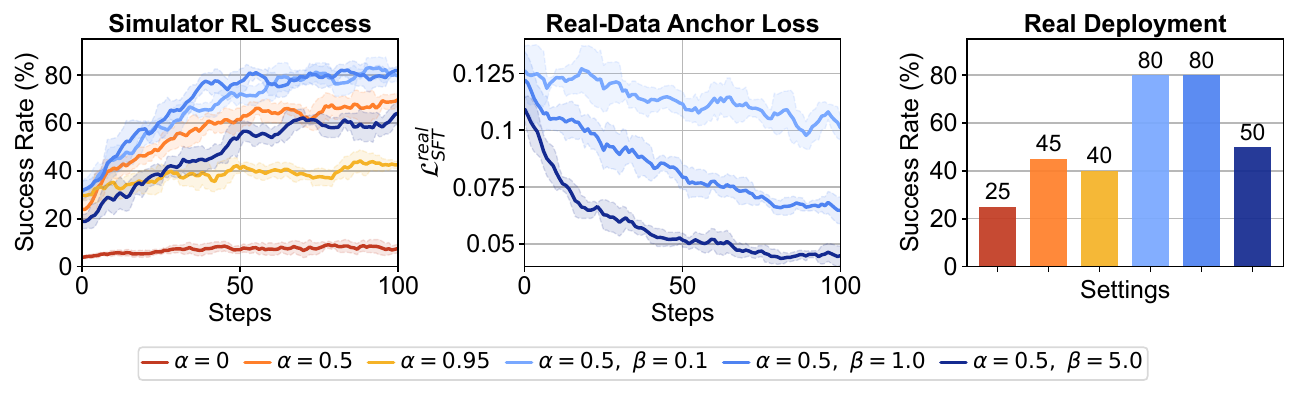}
  \vspace{-1.1em}
  \caption{\textbf{Ablation of co-training and real-data anchoring.} 
  We vary the co-training ratio $\alpha$ and real-anchor weight $\beta$ on the \emph{Assembly~\#2} task and report model performance in terms of simulator success rate \emph{(left)}, real-data anchor loss \emph{(middle)}, which measures policy deviation from real demonstrations during simulation, and real-world deployment success rate \emph{(right)}.
  }
  \label{fig:alpha_beta_ablation}
  \vspace{-0.5em}
\end{figure}

We further use the \emph{Assembly~\#2} task to ablate the two hyperparameters that govern sim-to-real co-training and simulator RL: the co-training ratio $\alpha$ sets the relative weight of simulated demonstrations during co-training, while the anchor weight $\beta$ determines the strength of the real-data anchor during RL. We report the simulator success rate, real-data anchor loss, and real-world deployment success rate across varying parameter settings in Figure~\ref{fig:alpha_beta_ablation}.

First, we study different choices of $\alpha$ with $\beta=0$. We observe that lowering $\alpha$ from $0.95$ to $0.5$ improves the best simulator success rate from $42.9\%$ to $70.3\%$ and increases real-world deployment success from $40\%$ to $45\%$, indicating that RL benefits from a more balanced sim-to-real co-training initialization. When $\alpha=0$, simulated demonstrations are removed entirely, so RL starts from a real-only policy in a zero-shot simulation setting. This configuration performs poorly in both domains ($14.1\%$ best simulator success and $25\%$ real success), suggesting that the initial simulator rollouts are too far from the policy's learned interaction distribution for effective post-training.

We then study the effect of $\beta$ with $\alpha=0.5$ fixed. Both $\beta=0.1$ and $\beta=1.0$ increase real-world deployment success from $45\%$ to $80\%$ and improve the best simulator success to above $92\%$. This indicates that the real-data anchor provides a useful balance: it prevents simulator RL from drifting away from the real-robot data distribution while still allowing the policy to learn reward-driven contact corrections. Without the anchor, the policy can overfit to simulator-specific contact strategies. In contrast, an overly strong anchor ($\beta=5.0$) yields the lowest supervised loss but suppresses RL refinement, pulling the policy back toward imitation and reducing both simulator and real-world success.

\vspace{-0.9em}
\section{Conclusion}

We presented \textbf{TacCoRL}, a sim-to-real post-training framework that augments pretrained VLA policies with tactile feedback. By combining sim-real co-training, simulator RL, and a real-data anchor, our framework learns contact-conditioned corrective actions that transfer to the real robot. Across four bimanual contact-rich tasks, our policy improves real-world success over vision-only and imitation-only baselines, highlighting a practical path for integrating tactile feedback into VLA policies via simulation.

\vspace{-0.8em}
\section{Limitations}

\noindent\textit{Requirement for real tactile data.}
Our method reduces the need for large-scale real tactile interaction data, but still uses a small set of real tactile demonstrations as an anchor during post-training. Future work could combine our approach with tactile representation learning and tactile simulation to further reduce the need for real tactile data collection.
\vspace{-0.5em}

\noindent\textit{Simulation setup effort.}
The current system uses a task-level digital twin; this is manageable for our setups but still involves manual effort. More automated asset reconstruction, camera alignment, and tactile-system identification would make the pipeline easier to scale.
\vspace{-0.5em}

\noindent\textit{Hard-to-simulate contacts.}
Our experiments focus on rigid or near-rigid contact-rich manipulation where the simulator can provide reliable contact predicates and tactile statistics. Tasks involving highly deformable objects or fluids require additional modeling beyond the current setup. Extending our framework to these more complex contact dynamics is an important direction for future work.

\bibliography{Corl26_ref}

\newpage
\appendix
\begin{center}
    \LARGE \bf Supplementary Materials
\end{center}
\addtocontents{toc}{\protect\setcounter{tocdepth}{2}}
\tableofcontents
\section{Robot Setup}
\label{app:robot_setup}
\begin{figure}[H]
    \centering
    \includegraphics[width=\linewidth]{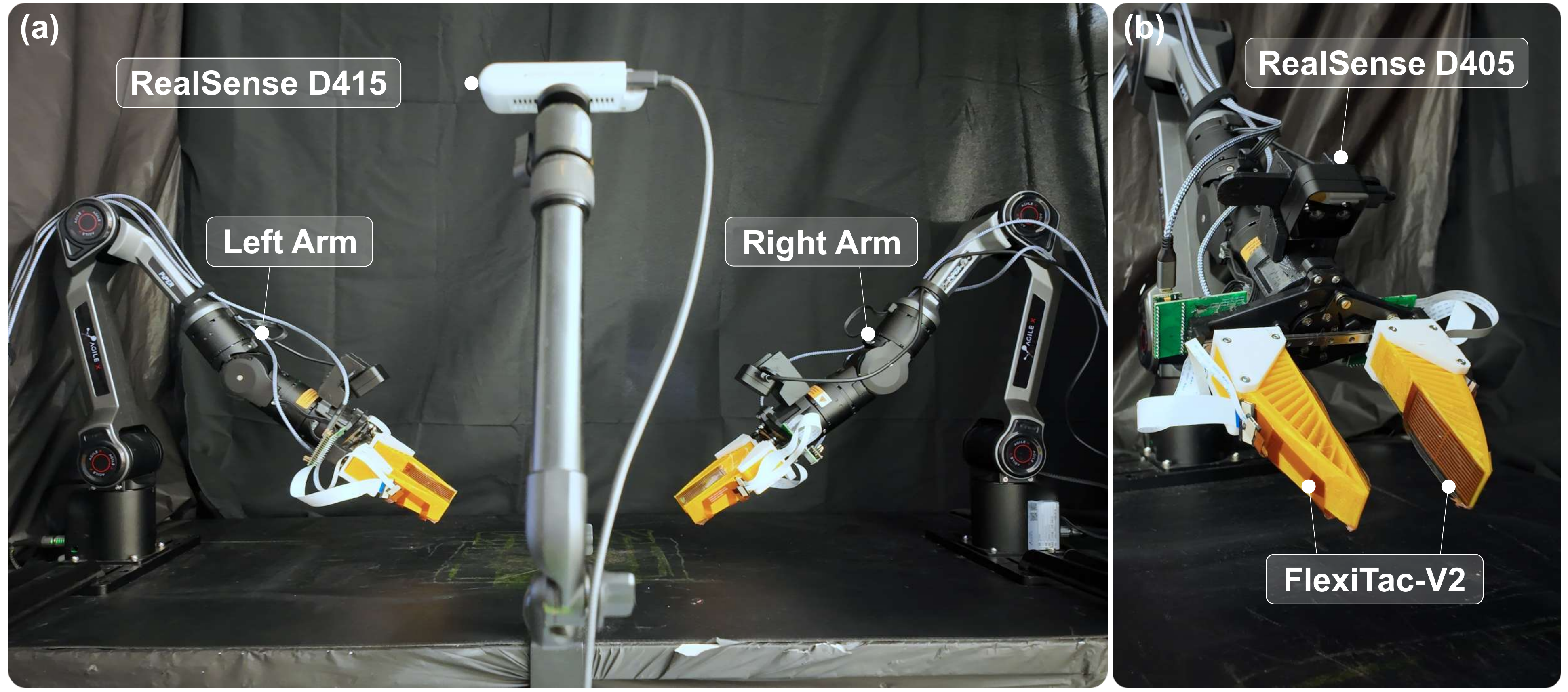}
    \caption{
    \textbf{Real-world robot setup.}
    (a) Bimanual platform with two AgileX PiPER 6-DoF robotic arms and a fixed front-view RealSense D415 camera.
    (b) End-effector close-up with wrist-mounted RealSense D405 cameras and two FlexiTac-V2 tactile pads on the gripper contact surfaces.
    }
    \label{fig:robot_setup}
\end{figure}

We conduct all real-world experiments on the bimanual platform in Figure~\ref{fig:robot_setup}. The setup uses two AgileX PiPER 6-DoF robotic arms, a fixed front-view RealSense D415 camera, two wrist-mounted RealSense D405 cameras, and two FlexiTac-V2 tactile pads \citep{huang2026flexitac}, each pad with a $32\times12$ taxel array, mounted on the opposing gripper contact surfaces.

\newpage
\section{Real-to-Sim-to-Real}
\label{app:real_to_sim_to_real}

\subsection{Controller SysID Details}
\label{app:controller_sysid}

\begin{figure}[H]
    \centering
    \includegraphics[width=\linewidth]{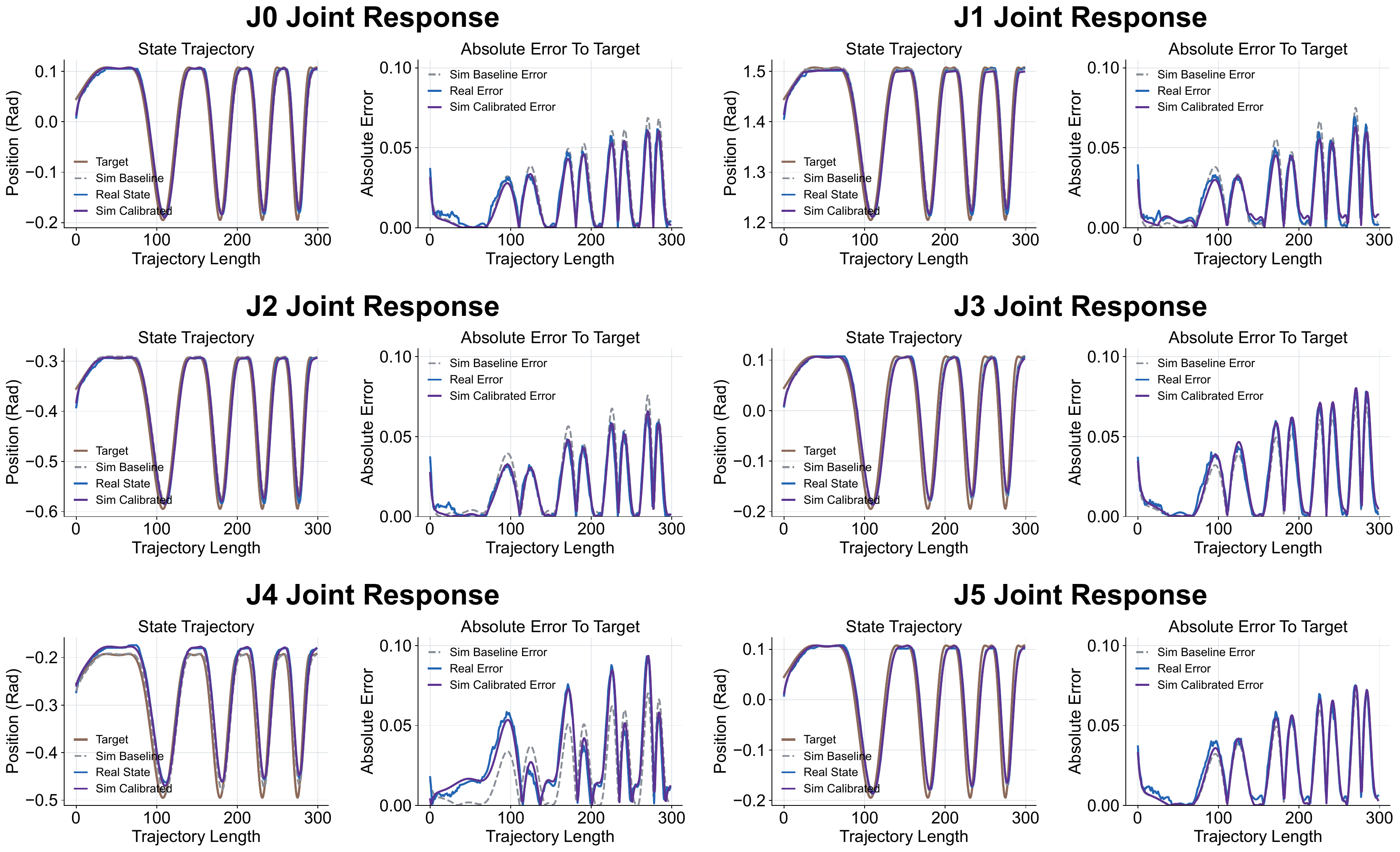}
    \small
    \setlength{\tabcolsep}{6pt}
    \begin{tabular}{lccc}
        \toprule
        Joint & $K_p$ ($\mathrm{N}\cdot\mathrm{m}\cdot\mathrm{rad}^{-1}$) & $K_d$ ($\mathrm{N}\cdot\mathrm{m}\cdot\mathrm{s}\cdot\mathrm{rad}^{-1}$) & $T_{\mathrm{ref}}$ ($\mathrm{N}\cdot\mathrm{m}$) \\
        \midrule
        J0 & 1813.0 & 159.4 & 0.12 \\
        J1 & 1392.4 & 123.7 & -0.44 \\
        J2 & 974.8 & 87.9 & 0.04 \\
        J3 & 587.1 & 69.8 & -0.12 \\
        J4 & 151.1 & 17.1 & 1.73 \\
        J5 & 151.8 & 16.5 & 0.07 \\
        \bottomrule
    \end{tabular}
    \caption{
    \textbf{Per-joint controller SysID.}
    Each panel replays the same real single-joint sweep in simulation.
    The reference simulator uses $K_p=500~\mathrm{N}\cdot\mathrm{m}\cdot\mathrm{rad}^{-1}$, $K_d=50~\mathrm{N}\cdot\mathrm{m}\cdot\mathrm{s}\cdot\mathrm{rad}^{-1}$, and $T_{\mathrm{ref}}=0~\mathrm{N}\cdot\mathrm{m}$; the calibrated simulator uses the identified parameters listed below the response plots.
    SysID reduces lag, overshoot, and steady-state bias across joints.
    }
    \label{fig:appendix_joint_response}
\end{figure}

For each joint $j$, we record a 10-second frequency sweep while holding the other joints fixed, then replay the same target sequence in simulation. We tune the simulated MIT-style impedance controller \citep{hogan1984impedance,katz2019mini} by optimizing $\boldsymbol{\eta}_j=[\log K_{p,j},\log K_{d,j},T_{\mathrm{ref},j}]$ with SPSA \citep{spall2002multivariate}, using the loss
\begin{equation}
    \mathcal{L}_j
    =
    \sum_t
    \left|
    e^{\mathrm{sim}}_{t,j}-e^{\mathrm{real}}_{t,j}
    \right|
    +
    0.1
    \sum_t
    \left|
    \Delta e^{\mathrm{sim}}_{t,j}-\Delta e^{\mathrm{real}}_{t,j}
    \right|,
    \qquad
    e^\rho_{t,j}=q^\rho_{t,j}-q^{\mathrm{tar}}_{t,j},
    \quad
    \rho\in\{\mathrm{real},\mathrm{sim}\}.
    \label{eq:appendix_controller_loss}
\end{equation}
The first term matches target-tracking bias, while the finite-difference term constrains response speed and damping. The search bounds are $K_p\in[10,10^{5}]~\mathrm{N}\cdot\mathrm{m}\cdot\mathrm{rad}^{-1}$, $K_d\in[1,10^{3}]~\mathrm{N}\cdot\mathrm{m}\cdot\mathrm{s}\cdot\mathrm{rad}^{-1}$, and $T_{\mathrm{ref}}\in[-8,8]~\mathrm{N}\cdot\mathrm{m}$. The reported runs use SPSA with $a_{\mathrm{SPSA}}=0.15$, $c_{\mathrm{SPSA}}=0.08$, $A_{\mathrm{SPSA}}=20$, $\alpha_{\mathrm{SPSA}}=0.602$, $\gamma_{\mathrm{SPSA}}=0.101$, gradient clipping at $5$, and early stopping patience of $20$ iterations. Please refer to Figure \ref{fig:appendix_joint_response} for the full PD calibration results.

\newpage

\subsection{Tactile Signal Alignment Details}
\label{app:tactile_alignment}
\vspace{-0.5em}

Let $\mathbf{o}^\tau_t$ be the tactile arrays flattened into $K$ taxels with scalar readings $f_{t,k}$. Following spring-damper tactile simulators \citep{akinola2025tacsl,huang2025vt}, each taxel queries the signed distance to the contacting object and maps penetration depth $\delta_{t,k}\ge 0$ and penetration-rate $\dot{\delta}_{t,k}$ to a normal contact reading:
\vspace{-0.4em}
\begin{equation}
    f^{\mathrm{sim}}_{t,k}
    =
    \left[
    k_n \delta_{t,k}
    +
    k_d \dot{\delta}_{t,k}
    \right]_+,
    \qquad
    [z]_+=\max(z,0),
    \label{eq:tactile_penalty_model}
\end{equation}
where $k_n$ and $k_d$ are normal stiffness and damping. Because the policy consumes scalar taxel histories, we align this normal channel rather than reconstructing shear or full force fields.

\begin{figure}[H]
    \centering
    \includegraphics[
        width=1.0\linewidth,
    ]{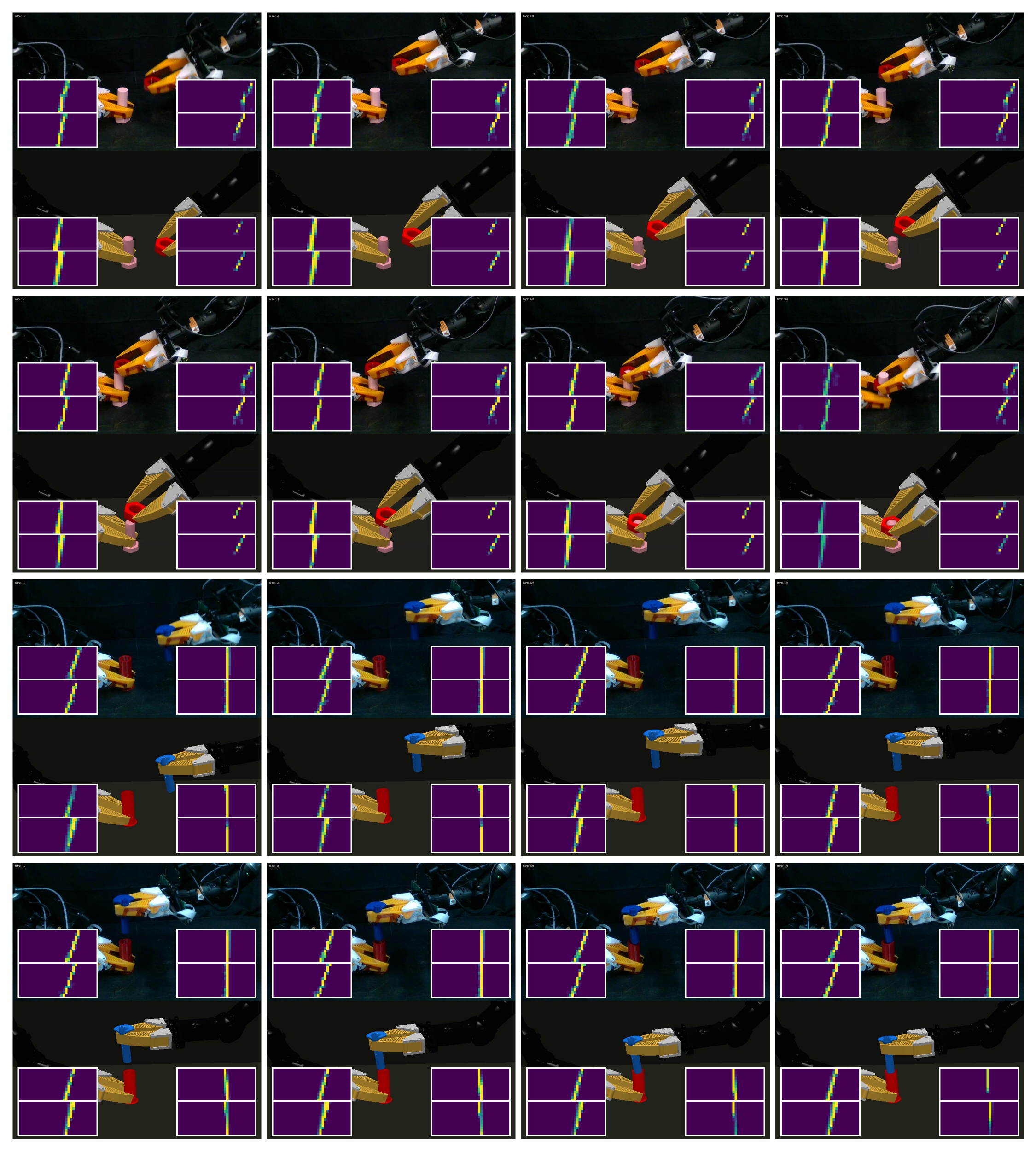}
    \vspace{-0.8em}
    \caption{
    \textbf{Qualitative tactile signal alignment.}
    Side-by-side real-to-sim replay trajectories for Assembly \#1 and \#2. Each pair shows synchronized real and simulated frames with tactile maps, highlighting matched contact location and evolution.
    }
    \label{fig:tactile_calibration}
\end{figure}

Alignment calibrates $(k_n,k_d)$ and the policy-facing normalization. We estimate $\lambda_{\mathrm{noise}}$ from no-contact real readings to suppress background noise and define active-taxel masks, then tune $(k_n,k_d)$ on matched real-to-sim contact replays to align active-taxel masks, active-taxel counts, centroids, histograms, and temporal differences. The same thresholding and normalization are applied in both domains; Figure~\ref{fig:tactile_calibration} shows the resulting side-by-side \emph{Assembly \#1} and \emph{Assembly \#2} replays.
\vspace{-0.0em}

\newpage

\subsection{Camera Calibration Details}
\label{app:camera_calibration}

\begin{figure}[H]
    \centering
    \includegraphics[width=\linewidth]{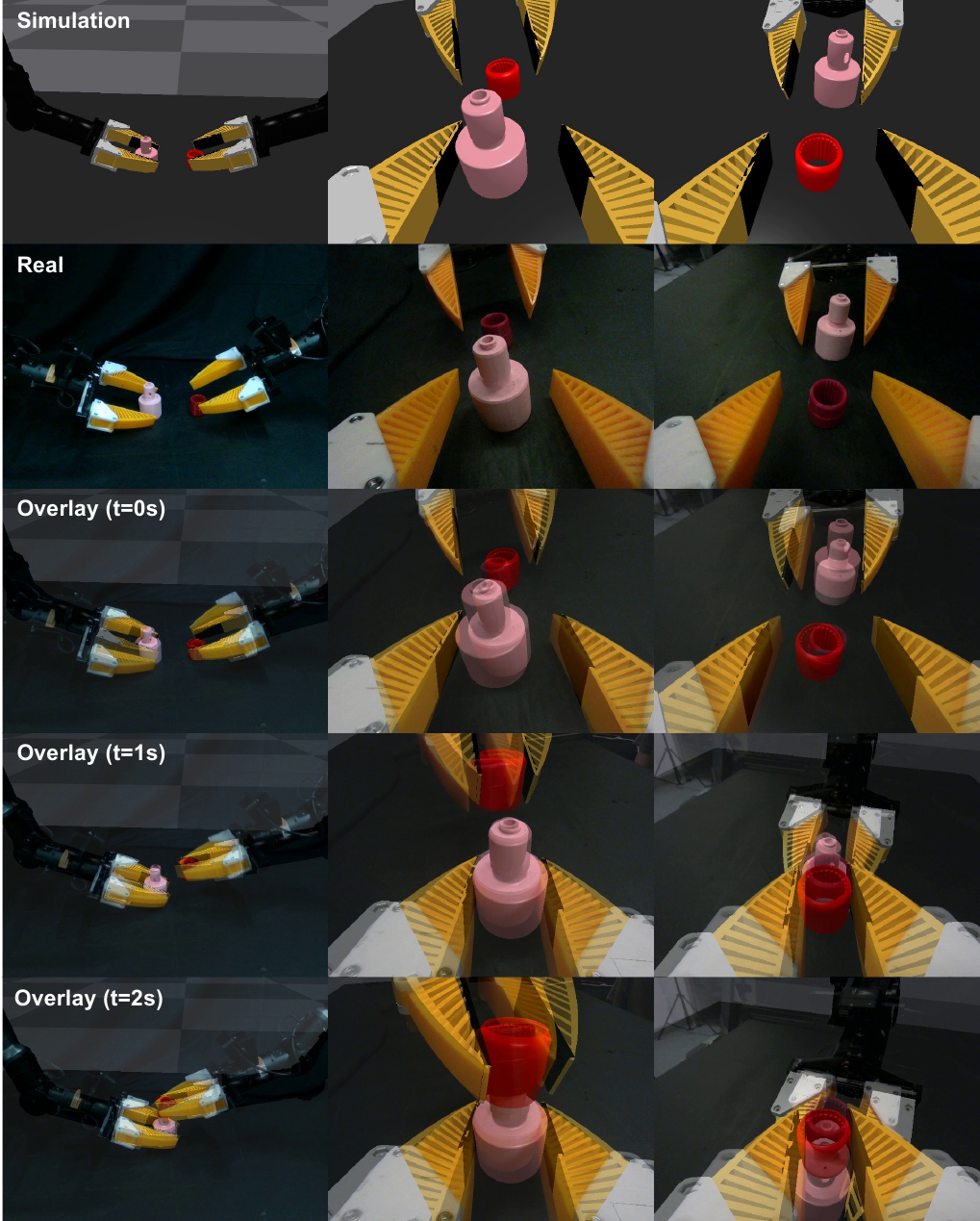}
    \caption{
    \textbf{Camera calibration.}
    Simulation-to-real alignment after camera extrinsic calibration.
    Columns show the fixed front camera, left wrist camera, and right wrist camera.
    The top two rows compare rendered simulation views with synchronized real views, and the bottom rows overlay the two domains at the initial state, grasp phase, and insertion phase.
    }
    \label{fig:appendix_camera_calibration}
\end{figure}

Camera extrinsics are calibrated offline by replaying a synchronized real episode in simulation and optimizing small pose offsets around the nominal camera mounts.
Figure~\ref{fig:appendix_camera_calibration} provides qualitative evidence of the resulting visual alignment across the initial, grasp, and insertion stages.

\newpage

\section{Baseline Failure Analysis}
\label{app:failure_analysis}

\begin{figure}[H]
    \centering
    \includegraphics[width=\linewidth]{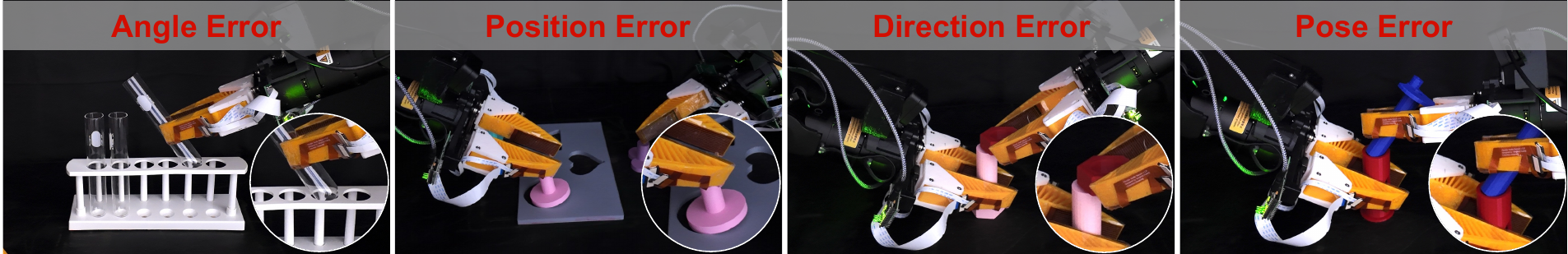}
    \vspace{-0.6em}
    \caption{
    \textbf{Representative baseline failure cases.}
    Each failure occurs after the gripper or held object occludes the local contact region. These cases expose missing corrective behavior: angle correction after rack-rim contact, lateral search for the puzzle opening, axis realignment in Assembly \#1, and center-of-mass support before release in Assembly \#2.
    }
    \label{fig:failureCases}
\end{figure}

\noindent Figure~\ref{fig:failureCases} shows one representative failure from each task; the full failure distribution is broader, but these examples share a post-contact ambiguity. After first contact, the rack rim, puzzle-hole boundary, or assembly mating interface is hidden by the gripper or held part, so success depends on converting local contact cues into corrective motion rather than continuing the nominal trajectory.

\noindent\textit{Test Tube Insertion:}
A small angular misalignment between the tube and rack opening causes the tube to catch on the rim. Once the gripper and tube hide the rack-rim contact line, vision cannot verify centering. The imitation-only tactile baseline detects contact but keeps inserting instead of reorienting, so the tube gets stuck.

\noindent\textit{Do Puzzle:}
A small lateral offset can leave the piece outside the target hole even when its orientation is roughly correct. The piece and gripper partially hide the hole boundary, and object symmetry makes the residual offset difficult to infer from vision alone. The imitation-only tactile baseline repeatedly presses at the same wrong location rather than translating to search, pushing the rollout into more out-of-distribution failure states.

\noindent\textit{Assembly \#1:}
An incorrect insertion direction after grasping can make the parts contact while the relative motion is not aligned with the true insertion axis. Once the held part and gripper hide the mating edge, the vision-only baseline continues the wrong approach, causing sliding or wedging.

\noindent\textit{Assembly \#2:}
Coupled translation and orientation error can leave the plug misaligned with the socket. Once the plug and gripper occlude the socket opening, the vision-only baseline may insert at an incorrect angle while the plug's center of mass remains outside the socket. After release, the plug is unsupported and falls out.

\newpage
\raggedbottom
\section{Training Parameters}
\label{app:cotrain_training_params}

We list training parameters for Secs.~\ref{sec:method_cotrain} and~\ref{sec:method_posttraining}. Table~\ref{tab:cotrain_training_params} covers fixed-dataset sim-real co-training for a tactile-augmented $\pi_{0.5}$ VLA model \citep{intelligence2025pi_}. For simulator post-training, the flow-policy RL implementation follows RLinf and $\pi_{\mathrm{RL}}$ \citep{yu2025rlinf,chen2025pirl}; Table~\ref{tab:posttraining_params} records the rollout, critic, PPO, and optimization settings.

\begin{table}[H]
    \caption{
    \textbf{Sim-real co-training parameter values.}
    Fixed-dataset sim-real co-training hyperparameters for a tactile-augmented $\pi_{0.5}$ VLA model \citep{intelligence2025pi_}.
    }
    \centering
    \footnotesize
    \setlength{\tabcolsep}{5pt}
    \renewcommand{\arraystretch}{1.01}
    \begin{tabular}{@{}>{\raggedright\arraybackslash}p{0.13\linewidth}>{\raggedright\arraybackslash}p{0.16\linewidth}>{\raggedright\arraybackslash}p{0.38\linewidth}>{\raggedright\arraybackslash}p{0.25\linewidth}@{}}
        \toprule
        Group & Symbol & Meaning & Value \\
        \midrule
        Data & $N_{\mathrm{real}}$ & real demonstration episodes & $50$ \\
         & $N_{\mathrm{sim}}$ & simulated demonstration episodes & $200$ \\
         & $\alpha$ & simulation loss weight & $0.5$ \\
        \addlinespace[0.5pt]\cmidrule(lr){2-4}\addlinespace[0.5pt]
        Tactile & $L$ & tactile history length & $10$ \\
         & $K$ & policy-facing taxels & $4\times32\times12=1536$ \\
         & $E_{\tau}$ & tactile encoder & CNN2 temporal token \\
         & $S_{\tau}$ & tactile injection site & prefix + suffix tokens \\
         & $M_{\mathrm{pre}}$ & prefix tactile tokens & $8$ \\
         & $M_{\mathrm{suf}}$ & suffix tactile tokens & $8$ \\
         & $\lambda_f$ & contact gate threshold & $0.4$ \\
         & $m$ & active-count threshold & $1$ \\
        \addlinespace[0.5pt]\cmidrule(lr){2-4}\addlinespace[0.5pt]
        Model & $H$ & action horizon & $10$ \\
         & $d_a$ & model output action dimension & $14$ \\
         & $\boldsymbol{\theta}_0$ & initialization & $\pi_{0.5}$ base VLA \\
        \addlinespace[0.5pt]\cmidrule(lr){2-4}\addlinespace[0.5pt]
        Training & $\sigma_v$ & image noise standard deviation & $0.03$ \\
         & $B$ & global batch size & $192$ \\
         & $T_{\mathrm{warm}}$ & warmup steps & $100$ \\
         & $\eta_{\max}$ & peak learning rate & $2.5{\times}10^{-5}$ \\
         & $\eta_{\min}$ & final learning rate & $5{\times}10^{-6}$ \\
         & $T_{\mathrm{co}}$ & co-training steps & \begin{tabular}[t]{@{}l@{}}{\tiny Assembly \#1/\#2:} $1000$\\{\tiny Do Puzzle/Test Tube Insertion:} $3000$\end{tabular} \\
         & $\mathcal{O}$ & optimizer & AdamW \\
        \bottomrule
    \end{tabular}
    \vspace{0.25em}
    \label{tab:cotrain_training_params}
\end{table}

\vspace{-0.6em}

\begin{table}[H]
    \caption{
    \textbf{Post-training parameter values.}
    Simulator post-training hyperparameters for Sec.~\ref{sec:method_posttraining}, including flow-policy RL, rollout, critic, PPO, and optimization settings.
    }
    \centering
    \footnotesize
    \setlength{\tabcolsep}{5pt}
    \renewcommand{\arraystretch}{1.03}
    \begin{tabular}{@{}>{\raggedright\arraybackslash}p{0.13\linewidth}>{\raggedright\arraybackslash}p{0.16\linewidth}>{\raggedright\arraybackslash}p{0.38\linewidth}>{\raggedright\arraybackslash}p{0.25\linewidth}@{}}
        \toprule
        Group & Symbol & Meaning & Value \\
        \midrule
        Anchor data & $N_{\mathrm{real}}^{\mathrm{anchor}}$ & real-anchor episodes & $200$ \\
         & $B_{\mathrm{real}}$ & real-anchor batch size & $32$ \\
         & $\beta$ & real-anchor loss weight & $1.0$ \\
        \addlinespace[0.5pt]\cmidrule(lr){2-4}\addlinespace[0.5pt]
        Rollout & $N_{\mathrm{env}}$ & parallel simulator environments & $128$ \\
         & $C$ & rollout chunk steps & $36$ \\
        \addlinespace[0.5pt]\cmidrule(lr){2-4}\addlinespace[0.5pt]
        Flow policy & $\mathcal{M}_{\mathrm{flow}}$ & flow-policy RL formulation & Flow-SDE \\
         & $\sigma_{\mathrm{SDE}}$ & Flow-SDE noise level & $0.5$ \\
         & $K_{\mathrm{denoise}}$ & denoising steps & $4$ \\
         & $\rho_{\mathrm{chunk}}$ & reward and log-prob aggregation & chunk level \\
        \addlinespace[0.5pt]\cmidrule(lr){2-4}\addlinespace[0.5pt]
        PPO & $\mathcal{A}$ & advantage estimator & GAE \\
         & $V_{\phi}$ & critic design & mean-token value head \\
         & $\gamma$ & discount factor & $0.99$ \\
         & $\lambda_{\mathrm{GAE}}$ & GAE trace parameter & $0.95$ \\
         & $\epsilon_{\mathrm{clip}}$ & PPO clip ratio & $0.2$ \\
         & $E_{\mathrm{update}}$ & PPO update epochs & $1$ \\
        \addlinespace[0.5pt]\cmidrule(lr){2-4}\addlinespace[0.5pt]
        Optimization & $B_{\mathrm{RL}}$ & RL global batch size & $512$ \\
         & $\eta_{\pi}$ & actor learning rate & $5{\times}10^{-6}$ \\
         & $\eta_V$ & value-head learning rate & $1{\times}10^{-4}$ \\
        \bottomrule
    \end{tabular}
    \vspace{0.25em}
    \label{tab:posttraining_params}
\end{table}

\newpage
\section{Domain Randomization Parameters}
\label{app:domain_randomization}

Table~\ref{tab:domain_randomization_params} summarizes the domain randomization used in simulation for each task. We include image-space augmentation and task-level reset randomization for object pose and physical properties.

\begin{table}[H]
    \caption{
    \textbf{Domain randomization parameter values.}
    Simulation domain-randomization values used across the tasks in Sec.~\ref{sec:exp_setting}.
    }
    \centering
    \footnotesize
    \setlength{\tabcolsep}{4pt}
    \renewcommand{\arraystretch}{1.08}
    \begin{tabular}{@{}>{\raggedright\arraybackslash}p{0.16\linewidth}>{\raggedright\arraybackslash}p{0.24\linewidth}>{\raggedright\arraybackslash}p{0.10\linewidth}>{\raggedright\arraybackslash}p{0.42\linewidth}@{}}
        \toprule
        Task & Parameter & Probability & Distribution \\
        \midrule
        All tasks & camera crop-resize & $1.0$ & $f_{\mathrm{crop}}=0.95$ \\
         & camera rotation & $1.0$ & $\Delta\theta\sim\mathcal{U}([-5^{\circ},5^{\circ}])$ \\
         & camera brightness jitter & $1.0$ & $j_b\sim\mathcal{U}([0.7,1.3])$ \\
         & camera contrast jitter & $1.0$ & $j_c\sim\mathcal{U}([0.6,1.4])$ \\
         & camera saturation jitter & $1.0$ & $j_s\sim\mathcal{U}([0.5,1.5])$ \\
         & object friction & $1.0$ & $\mu\sim\mathcal{U}([0.3,0.7])$ \\
        \addlinespace[0.5pt]\cmidrule(lr){2-4}\addlinespace[0.5pt]
        Assembly \#1/\#2 & plug/socket base position & $1.0$ & $\Delta \mathbf{p}\sim\mathcal{U}([-0.03,0.03]\,\mathrm{m})$ \\
         & plug/socket mass & $1.0$ & $\Delta m\sim\mathcal{U}([-0.03,0.03]\,\mathrm{kg})$ \\
        \addlinespace[0.5pt]\cmidrule(lr){2-4}\addlinespace[0.5pt]
        Do Puzzle & tray planar position & $1.0$ & $\Delta x,\Delta y\sim\mathcal{U}([-0.005,0.005]\,\mathrm{m})$ \\
         & tray yaw & $1.0$ & $\Delta\psi\sim\mathcal{U}([-5^{\circ},5^{\circ}])$ \\
         & piece planar position & $1.0$ & $\Delta x,\Delta y\sim\mathcal{U}([-0.01,0.01]\,\mathrm{m})$ \\
         & piece yaw & $1.0$ & $\Delta\psi\sim\mathcal{U}([-10^{\circ},10^{\circ}])$ \\
         & circle reset support & $1.0$ & $\sim\mathcal{U}([-0.03,0.03]^2\,\mathrm{m})$ \\
         & heart reset support & $1.0$ & $\sim\mathcal{U}([-0.03,0.03]\times[-0.027,0.027]\,\mathrm{m})$ \\
         & triangle reset support & $1.0$ & $\sim\mathcal{U}([-0.03,0.03]\times[-0.027,0.027]\,\mathrm{m})$ \\
        \addlinespace[0.5pt]\cmidrule(lr){2-4}\addlinespace[0.5pt]
        \multirow{3}{*}{\shortstack[l]{Test Tube\\Insertion}} & tube planar position & $1.0$ & $\Delta x,\Delta y\sim\mathcal{U}([-0.02,0.02]\,\mathrm{m})$ \\
         & tube yaw & $1.0$ & $\Delta\psi\sim\mathcal{U}([-20^{\circ},20^{\circ}])$ \\
         & holder planar position & $1.0$ & $\Delta x,\Delta y\sim\mathcal{U}([-0.005,0.005]\,\mathrm{m})$ \\
        \bottomrule
    \end{tabular}
    \vspace{0.25em}
    \label{tab:domain_randomization_params}
\end{table}

\end{document}